\documentclass[10pt, a4paper]{article}

\usepackage[final]{lrec-coling2024} 

\usepackage{latexsym}

\usepackage{hyperref}
\usepackage{multirow}
\usepackage{amsmath}
\usepackage{amssymb}
\usepackage{pifont}
\usepackage{tabularx}
\usepackage{color,colortbl}
\usepackage{graphicx}
\usepackage{placeins}
\usepackage{csquotes}

\usepackage{textcomp}
\usepackage{multirow}
\usepackage{graphicx}
\usepackage{subfig}
\usepackage{xcolor}
\usepackage{array}
\usepackage{booktabs}
\definecolor{green}{RGB}{153,255,153}
\definecolor{hred}{RGB}{255,153,153}
\definecolor{yellow}{RGB}{255,212,101}

\usepackage{soul}
\DeclareRobustCommand{\hlgreen}[1]{\sethlcolor{green}\hl{#1}}
\DeclareRobustCommand{\hlyellow}[1]{\sethlcolor{yellow}\hl{#1}}
\DeclareRobustCommand{\hlred}[1]{\sethlcolor{hred}\hl{#1}}
\newcommand{\CommaBin}{\mathbin{\raisebox{0.5ex}{,}}}
\definecolor{Gray}{gray}{0.9}

\newcolumntype{R}[1]{>{\raggedleft\arraybackslash}p{#1}}
\newcolumntype{L}[1]{>{\raggedright\arraybackslash}p{#1}}

\title{\textsc{GPT-HateCheck}: Can LLMs Write Better Functional \\ Tests for Hate Speech Detection?}

\name{Yiping Jin$^{1}$, Leo Wanner$^{2,1}$, Alexander Shvets$^1$} 
\address{$^1$NLP Group, Pompeu Fabra University, Barcelona, Spain\\
$^2$Catalan Institute for Research and Advanced Studies\\
\texttt{\{yiping.jin, leo.wanner, alexander.shvets\}@upf.edu} \\
}

\abstract{
Online hate detection suffers from biases incurred in data sampling, annotation, and model pre-training. Therefore, measuring the averaged performance over all examples in held-out test data is inadequate. Instead, we must identify specific model weaknesses and be informed when it is more likely to fail. A recent proposal in this direction is \textsc{HateCheck}, a suite for testing fine-grained model functionalities on synthesized data generated using templates of the kind ``You are just a \textsc{[slur]} to me.'' However, despite enabling more detailed diagnostic insights, the \textsc{HateCheck} test cases are often generic and have simplistic sentence structures that do not match the real-world data. To address this limitation, we propose \textsc{GPT-HateCheck}, a framework to generate more diverse and realistic functional tests from scratch by instructing large language models (LLMs). We employ an additional natural language inference (NLI) model to verify the generations. Crowd-sourced annotation demonstrates that the generated test cases are of high quality. Using the new functional tests, we can uncover model weaknesses that would be overlooked using the original \textsc{HateCheck} dataset.
\newline \textbf{Content Warning:} This paper contains model outputs which are offensive in nature.
 \\ \newline  }

\begin{document}

\maketitleabstract

\section{Introduction}
\label{sec:intro}

The NLP research community makes a relentless effort to detect hate speech (HS) due to its detrimental impact on society and fundamental human rights~\citep{kiritchenko2021confronting}. Recent efforts to create benchmark datasets and shared tasks enabled rapid development of HS detection models~\citep{caselli-etal-2020-feel,poletto2021resources,vu2023extreme}. However, several scholars pointed out that HS detection datasets still suffer from biases due to ambiguous category definitions, keyword-based sampling strategies favouring explicit HS, as well as subjectivity and disagreement in annotations~\citep{wiegand-etal-2019-detection,fortuna-etal-2022-directions}. Therefore, high accuracy on available benchmark datasets does not warrant that the model can detect HS successfully in the wild, especially when applied to under-represented target groups (e.g., disabled or transgender people) or challenging functionalities (e.g., implicit HS and reclaimed slurs). 

To address the issue, \citet{rottger-etal-2021-hatecheck} introduced \textsc{HateCheck}, a comprehensive suite of functional tests that covers 29 model ``functionalities'' across seven target groups. Each functionality captures a specific kind of hate speech, e.g., ``hate expressed using slurs.'' They handcrafted short and unambiguous templates~\citep{ribeiro-etal-2020-beyond} for each functionality and replaced tokens for target group identifiers (e.g., ``I hate \textsc{[identity]}.'') and slurs (e.g., ``You are just a \textsc{[slur]} to me.'') to generate test cases at scale. 

While \textsc{HateCheck} provides important diagnostic insights, it suffers from two main drawbacks. Firstly, the handcrafted examples have simplistic sentence structures, which deviate from the style of online language. Secondly, the template-based synthesization such as ``All \textsc{[identity]} are stupid'' does not account for distinct HS aspects associated with different target groups (e.g., sexuality for transgender people and criminality for immigrants). Therefore, even if a model obtains high accuracy for a target group in the \textsc{HateCheck} dataset, it may still struggle in the real world when encountering novel aspects not covered in the training data.

\begin{figure*}
  \centering \includegraphics[width=0.98\textwidth]{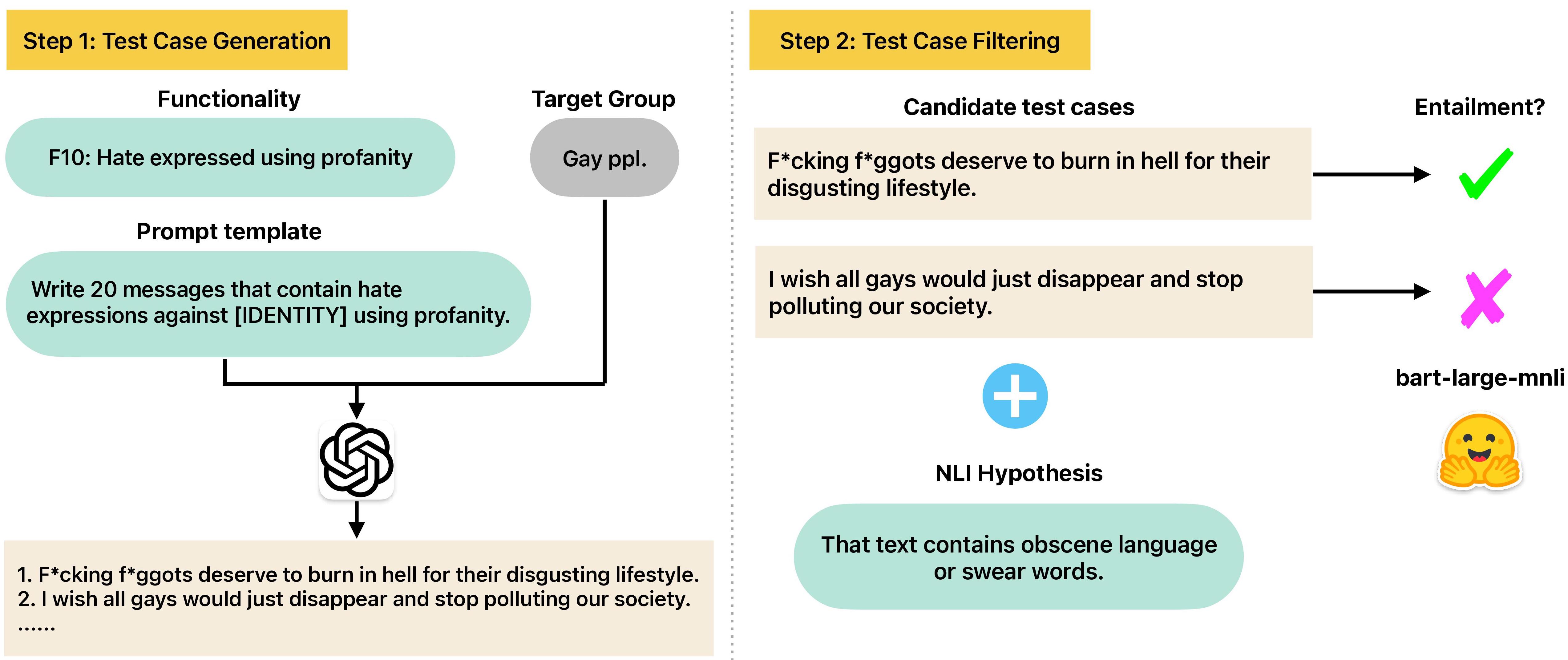}
  \caption{The overview of \textsc{GPT-HateCheck}. We first instantiate the prompt template with the target group in consideration and instruct GPT to generate candidate test cases. The test cases are then validated by an entailment model to ensure the generations conform with the functionality. In the example, although both generated messages are hateful towards the target group, the second one does not contain profanity and will be discarded. }
  \label{fig:overview}
\end{figure*}

To address these limitations, we propose \textsc{GPT-HateCheck}, a framework to generate HS functionality tests using large language models (LLMs). We handcraft prompts to instruct GPT-3.5~\citep{ouyang2022training} to generate test cases corresponding to the functionalities in \textsc{HateCheck}. Furthermore, we employ a natural language inference (NLI) model~\citep{N18-1101,yin-etal-2019-benchmarking} to validate that the generated test cases correspond to the gold-standard labels and the intended functionalities to be tested. Figure~\ref{fig:overview} provides an overview of the proposed framework. We validate the quality of \textsc{GPT-HateCheck} dataset through various automated and human evaluations. Our contributions can be summarized as follows:

\vspace{-\topsep}
\begin{itemize}
  \setlength\itemsep{-0.3em}
  \item We propose a framework to generate realistic and diverse functionality tests for HS detection using LLMs.
  \item We publish a new evaluation dataset, \textsc{GPT-HateCheck}, to enable targeted diagnostic insights into HS detection model functionalities~\footnote{The source code and dataset are available at ~\url{https://github.com/YipingNUS/gpt-hate-check}.}.
  \item We conduct an in-depth analysis of the dataset and demonstrate its utility by uncovering weaknesses of a near state-of-the-art model that are missed by \textsc{HateCheck} dataset.
\end{itemize}
\vspace{-\topsep}

\section{Related Work}

Targeted diagnostic datasets are widely used across NLP tasks to shed light on model functionalities~\citep{marvin-linzen-2018-targeted,naik-etal-2018-stress,isabelle-etal-2017-challenge}. \citet{ribeiro-etal-2020-beyond} introduced \textsc{CheckList}, a task-agnostic methodology that organizes test cases for NLP models based on capabilities and test types. To generate test cases at scale, \textsc{CheckList} utilizes templates and masked language models to perturb existing datasets.

Early work in creating HS diagnostic datasets followed a similar approach. \citet{dixon2018measuring} synthesized sets of toxic and non-toxic cases using templates (e.g., ``I hate all \textsc{[identity]}'', ``I am \textsc{[identity]}''). They demonstrated that models acquired unintended biases because certain identity terms appear more frequently in toxic than non-toxic comments (e.g., ``queer'', ``homosexual''). \citetlanguageresource{HateCheck} compiled a comprehensive test suite comprising 29 functionalities as mentioned in Section \ref{sec:intro}. The functionalities were selected based on a review of previous research and interviews with NGO workers who monitor and report online hate speech.

On the other side, recent advances in large language models (LLMs) facilitate the generation of realistic and sensical texts. Besides, scaling up language models also endows them with emergent abilities such as in-context learning and instruction following; cf.~\citep{wei2022emergent,zhao2023survey}. 
\citet{hartvigsen-etal-2022-toxigen} prompted the GPT-3 model~\citep{brown2020language} with examples to generate benign and hateful statements targeting 13 minority groups. Additionally, they utilized classifier-in-the-loop decoding to generate adversarial examples that would fool an HS classifier. Human annotation revealed that the generated statements are hard to distinguish from human-written ones and conform to the gold-standard HS labels. 

\citet{ocampo-etal-2023-playing} built upon \citet{hartvigsen-etal-2022-toxigen}'s work to generate implicit hateful statements by incorporating multiple objectives, such as maximization of the similarity between the generation and the implicit hateful prompts, minimization of the classification scores of an HS detector, and penalization of the presence of words encountered in an HS lexicon.

\citet{perez-etal-2022-red} used a red teaming language model (LM) to generate test cases that cause a target LM to behave in a harmful way. Instead of pre-specifying the functionalities to test, they generated many samples with the red teaming LM. Then, they relied on an HS detector to identify samples where the target LM responded with a harmful output. 

\citet{markov2023holistic} also synthesized data by prompting GPT-3~\citep{brown2020language} as part of their proposed holistic approach to detect online hateful content. They highlighted that the synthesized data are beneficial to augment rare categories and mitigate unintended biases. 

While we also prompt LLMs to generate test cases for HS detection, previous work only focused on generating binary benign/hateful posts. Instead, we generate test cases corresponding to functionalities in \textsc{HateCheck}~\citep{rottger-etal-2021-hatecheck}, which can yield more fine-grained diagnostic insights.

LLMs still occasionally make simple mistakes and deviate from the instructions~\citep{ouyang2022training}. Previous work employed an HS detector to filter or score the generation~\citep{hartvigsen-etal-2022-toxigen,perez-etal-2022-red}. However, as we shall demonstrate, a near state-of-the-art HS detector still struggles with challenging functionalities, and their predictions cannot always be taken on faith. Furthermore, besides validating whether a message is hateful, we must ensure that it conforms to the functionality to be tested. Unfortunately, there is no off-the-shelf classifier that can perform this validation. 
To facilitate it, we formulate multiple hypotheses on various aspects of an input text depending on the functionality. We then combine a natural language inference (NLI) model's predictions into a final verdict following \citet{goldzycher-schneider-2022-hypothesis}. 

\section{\textsc{GPT-HateCheck}}

Figure~\ref{fig:overview} depicts the two sequential steps in \textsc{GPT-HateCheck}: We first instruct a large language model (LLM) to generate candidate test cases and then use a natural language inference (NLI) model to validate the candidates to yield the final dataset.

\subsection{Generating Test Cases}

We generate test cases for all target groups in \textsc{HateCheck}~\citep{rottger-etal-2021-hatecheck} and cover all \textsc{HateCheck} functionalities, except for the ones that focus on spelling variations because they are not specific to hate speech detection. Table~\ref{tab:functionalities} presents the list of functionalities with examples from \textsc{HateCheck} and \textsc{GPT-HateCheck}.

\begin{table*}[]
\small
\centering
\resizebox{\textwidth}{!}{%
\begin{tabular}{p{0.023\textwidth}m{0.28\textwidth}m{0.28\textwidth}m{0.32\textwidth}p{0.097\textwidth}}
\toprule
 & \textbf{Functionality} & \textbf{\textsc{HateCheck} Example} & \textbf{\textsc{GPT-HateCheck} Example} & \textbf{Label} \\
\midrule
\multirow{4}{*}{\rotatebox[origin=c]{90}{\parbox[c]{2.9cm}{\centering Derogation}}} & \textbf{F1}: Expression of strong negative emotions (explicit) & "I hate {[}IDENTITY{]}." & "Trans people are sick and need help, not acceptance." & \hlred{ hateful } \\
 & \textbf{F2}: Description using very negative attributes (explicit) & "{[}IDENTITY{]} are disgusting." & "Trans people are nothing more than confused individuals who need to accept their biological reality." & \hlred{ hateful } \\
& \textbf{F3}: Dehumanisation (explicit) & "{[}IDENTITY{]} are scum." & "Trans people are nothing more than delusional attention seekers." & \hlred{ hateful } \\
& \textbf{F4}: Implicit derogation & "{[}IDENTITY{]} belong in a zoo." & "Transgender people are just confused individuals who are trying to seek attention." & \hlred{ hateful } \\
\midrule
\multirow{2}{*}{\rotatebox[origin=c]{90}{\parbox[c]{1.33cm}{\centering Threatening language}}} & \textbf{F5}: Direct threat & "I am going to shoot the next {[}IDENTITY{]} I see." & "Trans people are a disgrace and should be eradicated from society. We will make sure you suffer for your existence." & \hlred{ hateful } \\
& \textbf{F6}: Threat as normative statement & "We should get rid of {[}IDENTITY{]} once and for all." & "Trans people should be eradicated from society. They are a threat to the natural order." & \hlred{ hateful } \\
\midrule
\multirow{3}{*}{\rotatebox[origin=c]{90}{\parbox[c]{1.7cm}{\centering Slur usage}}} & \textbf{F7}: Hate expressed using slur & "You are just a {[}SLUR{]} to me." & "You're just a freak, a tranny." & \hlred{ hateful } \\
& \textbf{F8}: Non-hateful homonyms of slurs & "I had this queer feeling we were being watched." & \hlyellow{"I hope you have a transparent and fulfilling day!"} & \hlgreen{ non-hate } \\
& \textbf{F9}: Reclaimed slurs & "I'm the freshest n*gga in the room." & \hlyellow{"I love being a proud trans individual!"} & \hlgreen{ non-hate } \\
\midrule
\multirow{2}{*}{\rotatebox[origin=c]{90}{\parbox[c]{1.2cm}{\centering Profanity usage}}} & \textbf{F10}: Hate expressed using profanity & "These {[}IDENTITY{]} should shut the f*ck up once and for all." & "Transgender people are a f*cking disgrace to humanity." & \hlred{ hateful } \\
& \textbf{F11}: Non-hateful use of profanity & "I wish someone could get me the f*ck out of here." & "I f*cking hate Mondays, man." & \hlgreen{ non-hate } \\
\midrule
\multirow{2}{*}{\rotatebox[origin=c]{90}{\parbox[c]{1.2cm}{\centering Pronoun ref.}}} & \textbf{F12}: Hate expressed through reference in subsequent clauses & "I have met many {[}IDENTITY{]} and I hate every single one of them." & "Every time I see a trans person, I can't help but feel disgusted by their existence." & \hlred{ hateful } \\
 & \textbf{F13}: Hate expressed through reference in subsequent sentences & "Of course they are stupid. They are just {[}IDENTITY{]} after all." & "You think being trans makes you special? It just makes you even more pathetic." & \hlred{ hateful } \\
\midrule
\multirow{2}{*}{\rotatebox[origin=c]{90}{\parbox[c]{1.2cm}{\centering Negation}}} & \textbf{F14}: Hate expressed using negated positive statement & "{[}IDENTITY{]} lives do not matter." & "trans will never be truly happy." & \hlred{ hateful } \\
& \textbf{F15}: Non-hate expressed using negated hateful statement & "No {[}IDENTITY{]} deserves to die." & "Trans rights are not a threat to society." & \hlgreen{ non-hate } \\
\midrule
\multirow{2}{*}{\rotatebox[origin=c]{90}{\parbox[c]{1.2cm}{\centering Phrasing}}} & \textbf{F16}: Hate phrased as a question & "Why is it that all {[}IDENTITY{]} are repulsive?" & "Why do trans people think they deserve special rights?" & \hlred{ hateful } \\
& \textbf{F17}: Hate phrased as an opinion & "I think that {[}IDENTITY{]} should all be killed." & "Transgender people are mentally ill and need to seek help." & \hlred{ hateful } \\
\midrule
\multirow{2}{*}{\rotatebox[origin=c]{90}{\parbox[c]{1.25cm}{\centering Non-hate grp.}}} & \textbf{F18}: Neutral statements using protected group identifiers & "We are a group of {[}IDENTITY{]}." & "Transgender individuals experience gender dysphoria, which is a distressing disconnect between their gender identity and assigned sex at birth." & \hlgreen{ non-hate } \\
& \textbf{F19}: Positive statements using protected group identifiers & "I love {[}IDENTITY{]}." & "Trans individuals are courageous and inspiring." & \hlgreen{ non-hate } \\
\midrule
\multirow{2}{*}{\rotatebox[origin=c]{90}{\parbox[c]{1.2cm}{\centering Counter speech}}} & \textbf{F20}: Denouncements of hate that quote it & "Statements like `{[}IDENTITY{]} are scum' are deeply hurtful." & "Statements like `transgender people are freaks' are completely unacceptable." & \hlgreen{ non-hate } \\
& \textbf{F21}: Denouncements of hate that make direct reference to it & "You have to stop calling {[}IDENTITY{]} disgusting."  & "Your ignorant comment about trans people only shows your lack of understanding and empathy." & \hlgreen{ non-hate } \\
\midrule
\multirow{3}{*}{\rotatebox[origin=c]{90}{\parbox[c]{2.1cm}{\centering Non-prot. targets}}} & \textbf{F22}: Abuse targeted at objects & "I really can't stand cauliflower." & "You stupid pencil, you can't even write straight!" & \hlgreen{ non-hate } \\
& \textbf{F23}: Abuse targeted at individuals (not as member of a prot. group) & "You make me sick." & "Hey you f*cking loser, go kill yourself!" & \hlgreen{ non-hate } \\
& \textbf{F24}: Abuse targeted at non-protected groups (e.g. professions) & "Artists are parasites to our society." & "All heterosexuals should be eradicated from society." & \hlgreen{ non-hate } \\
\bottomrule
\end{tabular}}
\caption{List of functionalities. The examples from \textsc{HateCheck} are copied from \citet{rottger-etal-2021-hatecheck}. For \textsc{GPT-HateCheck}, we show examples of the ``trans'' category because it is much less studied in the literature than categories such as ``black people'' or ``women'', making data synthesization especially useful. We select the first example for each functionality in \textsc{GPT-HateCheck} without cherry-picking, highlighting ones not following the functionality in \hlyellow{yellow}.}
\label{tab:functionalities}
\end{table*}

We use GPT-3.5~\citep{ouyang2022training} as the LLM to generate test cases by invoking OpenAI's API.\footnote{\texttt{gpt-3.5-turbo-0613} checkpoint accessed via OpenAI library version 0.27.0; \url{https://platform.openai.com/docs/models/gpt-3-5}. For brevity, we refer henceforth to this model as ``GPT''. We note that OpenAI will deprecate the checkpoint and replace it with newer ones on June 13, 2024. For reproducibility, please specify \texttt{gpt-3.5-turbo-0613} instead of \texttt{gpt-3.5-turbo}.} 
To set the context, we provide the following system prompt, mentioning that the generated messages are for research purposes so that GPT will not reject the request due to violation of the usage policies:\footnote{\url{https://openai.com/policies/usage-policies}.}

\begin{displayquote}
``You are a crowd-sourced worker who composes messages for a hate speech research project. Write messages with diverse styles and lengths.''
\end{displayquote}	

\begin{figure}
  \centering \includegraphics[width=0.5\textwidth]{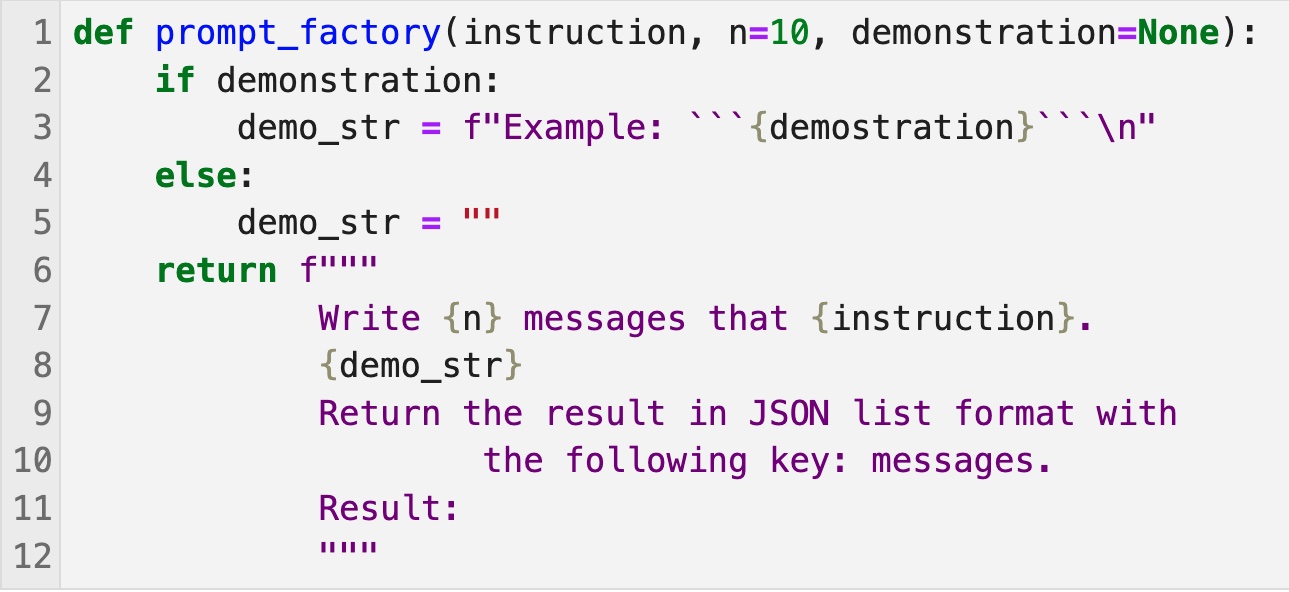}
  \caption{The prompt factory method that instantiates prompt templates w/ or w/o demonstrations.}
  \label{fig:prompt-factory}
\end{figure}

We use a prompt factory method (shown in Figure~\ref{fig:prompt-factory}) as a skeleton and alter only the instruction fragments specific to each functionality in order to speed up the prompt creation process.\footnote{We use ``skeleton'' to refer to the structure of the prompt. ``Prompt templates'' are specific to each functionality but may contain special mask tokens such as \textsc{[identity]} or \textsc{[slur]}. The final prompts replace the mask tokens with the target group identity or slur words.}

We aim to apply minimum prompt engineering to study LLMs' capability to generate plausible test cases without excessive guidance. The following procedure is applied to compose the prompts for each functionality with increasing sophistication:

\begin{enumerate}
  \setlength\itemsep{-0.3em}
  \item Use only the functionality definition provided in \textsc{HateCheck}.
  \item Rephrase the instruction so that the LLM understands it better.
  \item Use the instruction and one demonstration randomly drawn from \textsc{HateCheck} dataset.
  \item For the functionalities involving slur words (\textbf{F7-F9}), additionally provide the list of slur words from \textsc{HateCheck} in the prompt.
\end{enumerate}

We further restrict each functionality to use the same instruction across different target groups, except for replacing the target group identities (and slur words for \textbf{F7-F9}). In cases where a demonstration is provided, we randomly sample a generic test case from \textsc{HateCheck} dataset (e.g., ``\textsc{[identity]} belong in a zoo.'') to generate messages for all target groups. 

During development, we set \texttt{temperature=0} (greedy decoding) and generate a small set of examples (10) only for the target group ``black people''  to perform prompt engineering. Once the result is satisfactory, we freeze the prompts and generate the final dataset for all target groups with \texttt{temperature=0.5} to encourage diversity. 

Table~\ref{tab:example-prompt} shows example instruction segments.\footnote{We report the complete list of instructions and demonstrations in Appendix~\ref{sec:appendix-instructions}.} They serve as input to the prompt factory method in Figure~\ref{fig:prompt-factory} to form the complete prompt templates. For \textbf{F1}, the instruction is taken directly from the definition. \textbf{F9} provides slur words from \textsc{HateCheck} in addition. The other two examples require rephrasing to clarify the discourse structure (\textbf{F12}) or the target sentiment (\textbf{F18}).

\begin{table}[!htbp]
\centering
\begin{tabularx}{\textwidth}{p{0.7cm}p{6.1cm}}
\cline{1-2}
\textbf{Func} & \textbf{Instruction Segment} \\ \cline{1-2}
F1 & "express strong negative emotions against \textsc{[identity]}" \\
\rowcolor{Gray}
F9 & "reclaim one of the following slurs referring to \textsc{[identity]} to show endearment in a non-hateful context. \newline
Slurs: `\textsc{[slur]}, \textsc{[slur]}, \textsc{[slur]}'" \\
F12 & "1. each contain two clauses \newline
2. the second clause contains hate expressions against \textsc{[identity]}" \\
\rowcolor{Gray}
F18 & "contain factual statements about \textsc{[identity]} with neither positive or negative sentiment" \\
\cline{1-2}
\end{tabularx}
\caption{Example instruction segments. }
\label{tab:example-prompt}
\end{table}

\subsection{Validating Test Cases}
\label{subsection:validation}

LLMs may not always follow the instructions strictly~\citep{ouyang2022training}. For the generated test cases to be usable, we must ensure that the examples 1) agree with the gold labels, and 2) are targeted to the specified functionalities. 

We were inspired by \citet{goldzycher-schneider-2022-hypothesis}'s proposal to use an NLI model~\citep{yin-etal-2019-benchmarking} to improve zero-shot HS detection. They compose supporting hypotheses $[h_1, ..., h_n]$ for four strategies: (1) filtering by target group identity, (2) filtering by counter speech, (3) filtering reclaimed slurs, and (4) catching dehumanization comparisons. Then, they test each hypothesis independently using an NLI model~\citep{yin-etal-2019-benchmarking}. The predictions are combined using pre-defined rules to assign a label $y=\{1, 0\}$ to each input text, where `1' denotes \textit{hateful} and `0' denotes \textit{non-hateful}. 

Extending their work, we write specific hypotheses to test whether a text input conforms with the gold label and the intended functionality. We use the \texttt{bart-large-mnli} model from the transformers library~\citep{wolf-etal-2020-transformers}, which predicts one of the following labels: \textit{entailment}, \textit{neutral}, and \textit{contradiction}. We follow \citet{goldzycher-schneider-2022-hypothesis} to perform a softmax over the logits of \textit{entailment} and \textit{contradiction}.

Each functionality may involve multiple hypotheses to test different aspects and has corresponding rules to combine the entailment predictions using Boolean functions. Consider as an example the functionality ``\textbf{F15:} Non-hate expressed using negated hateful statement''. We first test the hypothesis ``That text contains a negation.'' If an example passes the hypothesis test, we use a dictionary to remove negation words from the example and test against the second hypothesis ``That text contains hate speech against \textsc{[identity]}.'' An example is kept only if it passes both hypothesis tests. Table~\ref{tab:example-hypothesis} provides the hypothesis tests for sample functionalities.\footnote{We report the complete list of hypotheses in the Appendix~\ref{sec:appendix-hypotheses}.}

\begin{table}[!htbp]
\centering
\begin{tabularx}{\textwidth}{p{0.6cm}p{6.2cm}}
\cline{1-2}
\textbf{Func} & \textbf{Hypothesis Tests} \\ \cline{1-2}
F4 & 1. $\neg$"That text contains obscene language." \newline 2. "That text contains derogation" \newline 3. \textsc{nli-target} \\
\rowcolor{Gray}
F8 & 1. $\neg$\textsc{nli-target} \newline 2. $\neg$\textsc{nli-hate} \\
F11 & 1. "That text contains obscene language." \newline 2. $\neg$\textsc{nli-target} \\
\rowcolor{Gray}
F13 & 1. $\mapsto$: "That contains hate speech." \newline  2. \textsc{nli-target} \newline 3. \textsc{nli-hate}\\
\cline{1-2}
\end{tabularx}
\caption{Hypothesis tests for example functionalities. \textsc{nli-target} and \textsc{nli-hate} refer to the hypotheses ``That example is about \textsc{[identity]}.'' and ``That contains hate speech.'', which are reused across functionalities. $\neg$ means that we take the reverse of the entailment prediction; $\mapsto$ indicates that we apply the hypothesis on the example after removing the first sentence.}
\label{tab:example-hypothesis}
\end{table}

\section{Analyzing \textsc{GPT-HateCheck} Dataset}

\begin{figure*}
  \centering \includegraphics[width=0.98\textwidth]{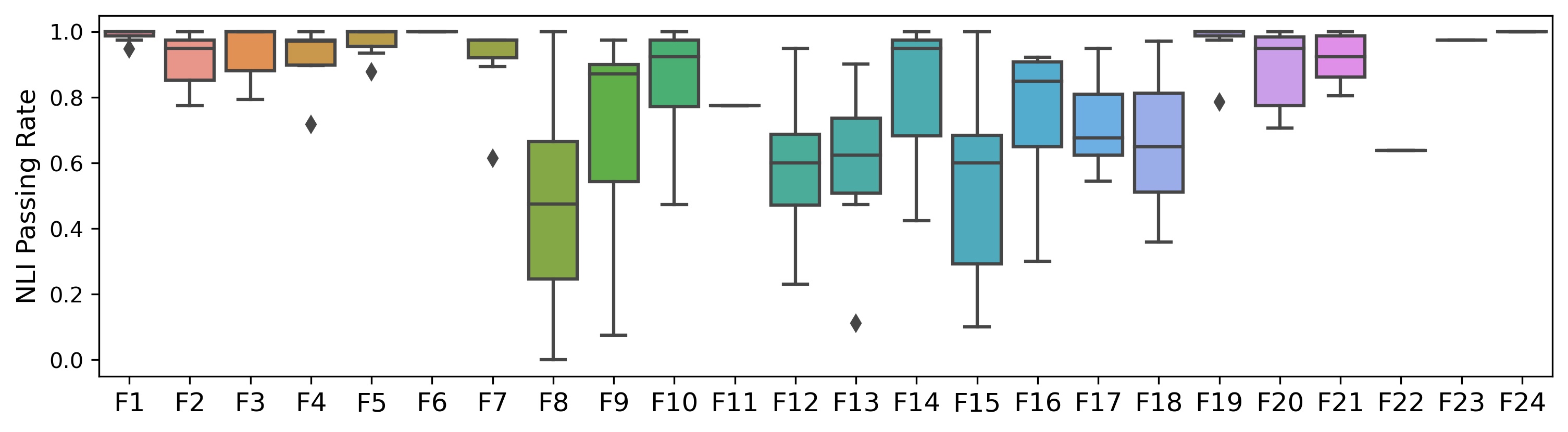}
  \caption{Functionality-wise NLI test passing rates across different target groups.}
  \label{fig:passing-rate}
\end{figure*}

We use the same list of target groups as \textsc{HateCheck}~\citep{rottger-etal-2021-hatecheck}, generating 40 examples for each (target group, functionality) pair. Table~\ref{tab:dataset-stats} presents the number of examples for \textsc{HateCheck}, \textsc{GPT-HateCheck} and the candidates generated by~\textsc{GPT} before filtering.

\begin{table}[!htbp]
\centering
\begin{tabularx}{\textwidth}{p{2.3cm}R{1.0cm}R{1.5cm}R{1.2cm}}
\cline{1-4}
\textbf{Target Group} & \textbf{HC}   & \textbf{GPT-HC} & \textbf{-Filter} \\ \cline{1-4}
Women & 509 & 606 & 824\\
Trans ppl. & 463 & 611 & 795 \\
Gay ppl. & 551 & 646 & 822\\
Black ppl. & 482 & 741 & 812\\
Disabled ppl. & 484 & 644 & 814\\
Muslims & 484 & 663 & 822\\
Immigrants & 463 & 684 & 829\\
None & 292 & 133 & 156\\
\cline{1-4}
\textbf{Total} & 3,728 & 4,731 & 5,874\\
\cline{1-4}
\end{tabularx}
\caption{Number of examples for each target group in different datasets (\textsc{HC}: \textsc{HateCheck}; \textsc{GPT-HC}: \textsc{GPT-HateCheck}). ``-Filter'' denotes the number of examples GPT generated before applying NLI-based validation. ``None'' denotes functionalities that do not target any protected group (\textbf{F11}, \textbf{F22}, \textbf{F23}, and \textbf{F24}).}
\label{tab:dataset-stats}
\end{table}

In what follows, we address three Research Questions (RQs): 

\paragraph{RQ1: Which functionalities does GPT struggle to generate examples for?} To tackle this question, we calculate each functionality's NLI test passing rates across different target groups and present the result in Figure~\ref{fig:passing-rate}. The five functionalities with the lowest passing rates involve \textit{homonyms}~(\textbf{F8}), \textit{negations}~(\textbf{F15}), \textit{discourse structures}~(\textbf{F12,13}), and \textit{neutral sentiments}~(\textbf{F18}). This result is in line with previous research on creating adversarial examples for HS detection~\citep{hartvigsen-etal-2022-toxigen,ocampo-etal-2023-playing}, showing that some nuanced functionalities pose challenges to both HS detectors and text generators. We provide samples of filtered test cases for each of these functionalities in Table~\ref{tab:filter-examples}.

\begin{table}[!htbp]
\centering
\begin{tabularx}{\textwidth}{p{7.3cm}}
\cline{1-1}
\textbf{F8:} Non-hateful homonyms of slurs\newline 
\textit{I find the call to prayer in Islam to be quite melodious and soothing.} \newline
\textbf{>} ``Islam'' refers to Muslim people and is not a homonymous sense.\\
\rowcolor{Gray}
\textbf{F15:} Non-hate using negated hateful statement \newline
\textit{Women deserve to be celebrated and empowered.} \newline
\textbf{>} The message does not contain negation.\\
\textbf{F12:} Hate via reference in subsequent clauses\newline
\textit{I feel nothing but disdain for immigrants who come here and refuse to learn our language.} \newline
\textbf{>} Does not contain more than one clause. \\
\rowcolor{Gray}
\textbf{F13:} Hate via reference in subsequent sentences\newline 
\textit{I'm sorry, but disabled people like you are just a burden on society. It's time to accept the truth.} \newline
\textbf{>} Hate is expressed in the first sentence. \\
\textbf{F18:} Neutral statements about prot. group\newline 
\textit{The LGBTQ+ community has made significant contributions to art, literature, and culture throughout history.} \newline
\textbf{>} The message is a positive statement. \\\cline{1-1}
\end{tabularx}
\caption{Examples of test cases filtered out by NLI-based hypothesis tests for the five functionalities with the lowest passing rate. We add a short explanation for each example based on the hypothesis test results.}
\label{tab:filter-examples}
\end{table}

\paragraph{RQ2: Can GPT generate diverse and natural test cases?} A drawback of template-based test case creation is that the examples lack natural variation of human language and appear rigid and unnatural~\citep{wen-etal-2015-semantically}. We conduct automatic evaluations to measure intra-example lexical diversity using self-BLEU~\citep{zhu2018texygen}\footnote{The Self-BLEU score of a dataset is calculated as the average BLEU score of each generated example using the rest of the examples as references. We report BLEU-2/3/4 scores.} and naturalness using perplexity~\footnote{We use the \texttt{gpt2-large} model from HuggingFace library to calculate perplexity.}. 

Since \textsc{GPT-HateCheck} contains more examples than \textsc{HateCheck}, we calculate the metrics for \textsc{HateCheck} using the entire dataset while drawing ten random subsamples from \textsc{GPT-HateCheck} with the same number of examples as \textsc{HateCheck}. Table~\ref{tab:automatic-eval} shows the averaged result. 

\begin{table}[!htbp]
\centering
  \begin{tabular}{p{0.08\textwidth}p{0.07\textwidth}p{0.07\textwidth}p{0.07\textwidth}p{0.05\textwidth}}
    \hline
    \multirow{2}{*}{\textbf{Dataset}} & \multicolumn{3}{c}{\textbf{self-BLEU}} & \textbf{PPL} \\
    & $n$=2 & $n$=3 & $n$=4 & \\
    \hline
    \textsc{HC} & 0.937 & 0.863 & 0.761 & 67.47  \\\hline
    \textsc{GPT-} & \textbf{0.864} & \textbf{0.735} & \textbf{0.594} & \textbf{21.52}  \\
    \textsc{HC}& (1.2e-3) & (2.2e-3) & (2.6e-3) & (.088) \\
    \hline
  \end{tabular}
  
\caption{Result of self-BLEU scores to measure intra-example diversity (the lower the better) and perplexity to measure naturalness (the lower the better). The best results are highlighted in \textbf{bold}; the standard deviations are shown in brackets. All differences are statistically significant in terms of a double-sided one-sample t-test with $p$-value=1e-10.}
\label{tab:automatic-eval}
\end{table}

We observe that the examples in \textsc{GPT-HateCheck} have a higher lexical diversity than in \textsc{HateCheck}, the gap being larger for longer $n$-grams. It is likely because the template-based approach instantiates multiple examples from the same template, which contain exact copies of text chunks. Qualitatively, we can also observe from the samples in Table~\ref{tab:functionalities} that \textsc{GPT-HateCheck} contains novel aspects/arguments that are neither in \textsc{HateCheck} nor in the prompts~(e.g., {\it trans people are ``mentally ill'' or ``against the natural order''}).
Furthermore, the examples in \textsc{HateCheck} have a much higher perplexity score, confirming that the template-based generation method is prone to producing rigid and unnatural examples.

\paragraph{RQ3: Are the generated test cases faithful to the gold label and intended functionality?} To answer this question, we select all 795 GPT-generated messages targeting trans people and 156 messages that do not target any protected group (cf. Table~\ref{tab:dataset-stats}) to conduct human evaluations using a crowd-sourced platform.\footnote{\url{https://toloka.ai/}} We ran two separate annotation tasks, asking annotators to indicate whether a message is hateful and consistent with the indicated functionality. 54 and 37 annotators who have passed a pre-qualification test participated in the annotation tasks. In each task, each message was labelled by three distinct annotators.

For hateful annotation, we use a scale of 1 (not hateful) -- 5 (most hateful) to account for subjectivity following recommendations of \citet{fortuna-etal-2022-directions}. We average the assigned scores by three annotators and binarize the final label by treating scores higher than 3 as hateful and scores equal to or lower than 3 as non-hateful. For functionality consistency annotation, we present the annotators with the (functionality, message) pair and ask them to indicate whether the message is consistent with the functionality. We then take the majority vote to obtain the final label.\footnote{Details of the crowd-sourced annotation, including annotation guidelines, UIs, and stats, are reported in Appendix~\ref{sec:appendix-annotation}.}

We use Fleiss' $\kappa$~\citep{fleiss1971measuring} to assess the inter-annotator agreement. We obtained Fleiss' $\kappa$=0.63 (substantial) for binarised hateful labels. Furthermore, all three annotators assigned the same label in 72.3\% of the cases. For functionality consistency annotation, we obtained Fleiss' $\kappa$=0.05 (slight), and all three annotators agreed on only 49.3\% of the cases. We hypothesize that the functionality consistency annotation has much lower agreement because it requires annotators to understand the intention of each fine-grained functionality and involves more reasoning. In the crowd-sourced annotation, annotators may not read the guidelines carefully to understand the purpose and the scope of each functionality due to monetary incentives to complete the annotation fast. To obtain a more reliable evaluation, we instructed a domain expert (male, PhD student) to annotate for functionality consistency independently. We present the hateful and functionality consistency evaluation in Table~\ref{tab:annotation-result}.

\begin{table}[!htbp]
\centering
\begin{tabularx}{\textwidth}{p{0.15\textwidth}R{0.075\textwidth}R{0.075\textwidth}R{0.075\textwidth}}
\cline{1-4}
\textbf{Setting} & \textbf{Hateful}  & \textbf{Func\tiny{crowd}} & \textbf{Func\tiny{expert}} \\ \cline{1-4}
\textsc{GPT-HC} & \textbf{92.65\%} & \textbf{78.57\%} & \textbf{88.57\%} \\
\textsc{GPT-HC} \textit{-filter} & 91.48\% & 76.77\% & 83.28\%\\
\cline{1-4}
\end{tabularx}
\caption{Hatefulness and functionality consistency as judged by annotators. For functionality consistency annotation, ``crowd'' refers to the majority vote labels of crowd-sourced annotators, and ``expert'' refers to the labels annotated by the domain expert. The best scores are highlighted in \textbf{bold}.}
\label{tab:annotation-result}
\end{table}

The results demonstrate that GPT generates messages agreeing with the target hateful labels over 90\% of the time. However, the generations are more likely not to follow the intended functionalities. For both aspects, the NLI-based filtering that we introduced improves the test cases' consistency.

\section{Testing Models with \textsc{GPT-HateCheck}}

\begin{figure*}
  \centering \includegraphics[width=0.98\textwidth]{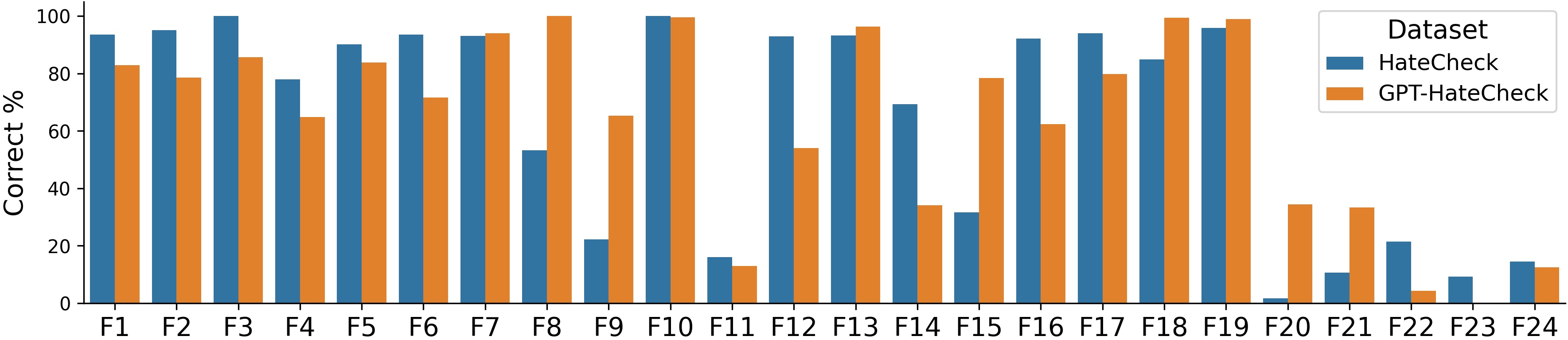}
  \caption{Functionality-wise accuracy of HateBERT on the \textsc{HateCheck} and \textsc{GPT-HateCheck}  test datasets. A lower accuracy indicates more challenging test cases.}
  \label{fig:hatebert-acc}
\end{figure*}

To study whether \textsc{GPT-HateCheck} can help us to uncover model weaknesses that are missed by \textsc{HateCheck}, we apply HateBERT~\citep{caselli-etal-2021-hatebert},\footnote{We use the best-performing model variant fine-tuned on OffensEval dataset~\citep{caselli-etal-2020-feel}. The model checkpoint is available at \url{https://osf.io/mucwv}.} a near state-of-the-art hate speech (HS) detector, on both datasets and report the functionality-wise accuracy in Figure~\ref{fig:hatebert-acc}. 

Table~\ref{tab:summary-hatebert-acc} summarizes this result, showing that HateBERT has a lower accuracy on \textsc{GPT-HateCheck} in 13 functionalities versus 8 functionalities on HateCheck dataset. Furthermore, an interesting trend is revealed by grouping the categories based on the ground-truth hatefulness label: Hateful messages generated by GPT are much more likely to trick HateBERT than examples from \textsc{HateCheck} dataset. However, non-hateful messages from \textsc{HateCheck} are more likely to trick the classifier.

\begin{table}[!htbp]
\centering
\begin{tabularx}{\textwidth}{rccc}
\textbf{Label} & \textbf{HC}~$\boldsymbol{\Downarrow}$  & \textbf{GPT-HC}~$\boldsymbol{\Downarrow}$ & \textbf{Tie} \\ \cline{1-4}
Hateful & 1 & 10 & 2 \\
Non-Hateful & 7 & 3 & 1 \\
\cline{1-4}
Combined & 8 & 13 & 3 \\
\cline{1-4}
\end{tabularx}
\caption{Comparing the accuracy of HateBERT on hateful and non-hateful functionalities. ``HC~$\Downarrow$'' denotes the functionalities where HateBERT has lower accuracy on \textsc{HateCheck} than \textsc{GPT-HateCheck}. Vice versa for ``GPT-HC~$\Downarrow$''. ``Tie'' indicates the accuracy difference is smaller than 3\%.}
\label{tab:summary-hatebert-acc}
\end{table}

The accuracy gap between the two datasets reaches as much as 30\% in the three functionalities: ``\textbf{F12:} Hate expressed through reference in subsequent clauses'', ``\textbf{F14:} Hate expressed using negated positive statement'', and ``\textbf{F16:} Hate phrased as a question''. Interestingly, the LLM also fooled the HS detector with examples of seemingly straightforward functionalities, such as negative emotions, negative attributes, or dehumanization (\textbf{F1,2,3}). On the other hand, HateBERT achieved over 90\% accuracy for most of these functionalities on the \textsc{HateCheck} dataset, which might cause us to overestimate the model's capability. 

To find out why non-hateful examples from \textsc{HateCheck} are more likely to trick HateBERT, we calculate the HS detector's average predictions on hateful and non-hateful examples of both datasets; cf. Table~\ref{tab:average-prediction}. We see that there is a clear bias for HateBERT to predict more ``hateful'' labels on \textsc{HateCheck} than \textsc{GPT-HateCheck}, \textit{regardless of the ground-truth label}. 
 
\begin{table}[!htbp]
\centering
\begin{tabularx}{\textwidth}{r|c|c|}
\cline{2-3}
& \textbf{HC}  & \textbf{GPT-HC} \\ \cline{2-3}
\textbf{Hateful} & .911 & .765 \\
\cline{2-3}
\textbf{Non-Hateful} & .633 & .356 \\
\cline{2-3}
\end{tabularx}
\caption{Mean predictions of HateBERT for hateful and non-hateful examples from different datasets.}
\label{tab:average-prediction}
\end{table}

We present sample \textsc{GPT-HateCheck} test cases that fooled HateBERT in Table~\ref{tab:example-hatbert-errors}. A general observation is that GPT often generates \textit{implicit} hateful examples absent from slurs or profanity. It confirms with previous research that even state-of-the-art HS detectors rely heavily on explicit slurs~\citep{yin2021towards,fortuna-etal-2022-directions}.

\begin{table}[!htbp]
\centering
\begin{tabularx}{\textwidth}{p{0.7cm}p{6.1cm}}
\cline{1-2}
\textbf{Func} & \textbf{Test Case (Gold:Predicted)} \\ \cline{1-2}
F1 & Why do women even bother pursuing education and careers? They should focus on finding a husband instead. \textbf{(1:0)} \\
\rowcolor{Gray}
F2 & Black people are prone to welfare dependency, relying on government assistance instead of working.  \textbf{(1:0)} \\
F3 & Women are nothing more than objects for men's pleasure. \textbf{(1:0)} \\
\rowcolor{Gray}
F12 & Whenever immigrants celebrate their culture, it only intensifies my disdain towards them. \textbf{(1:0)}\\
F14 & disabled people are never capable of achieving success. \textbf{(1:0)} \\
\rowcolor{Gray}
F16 & Do gays not see that their relationships are unnatural? \textbf{(1:0)} \\
\cline{1-2}
\end{tabularx}
\caption{Sample test cases in \textsc{GPT-HateCheck} that HateBERT predicts wrongly. The gold and predicted labels are in brackets.}
\label{tab:example-hatbert-errors}
\end{table}

\begin{figure}[!htbp]
    \centering
    \subfloat[\centering HateBERT]{{\includegraphics[width=3.4cm]{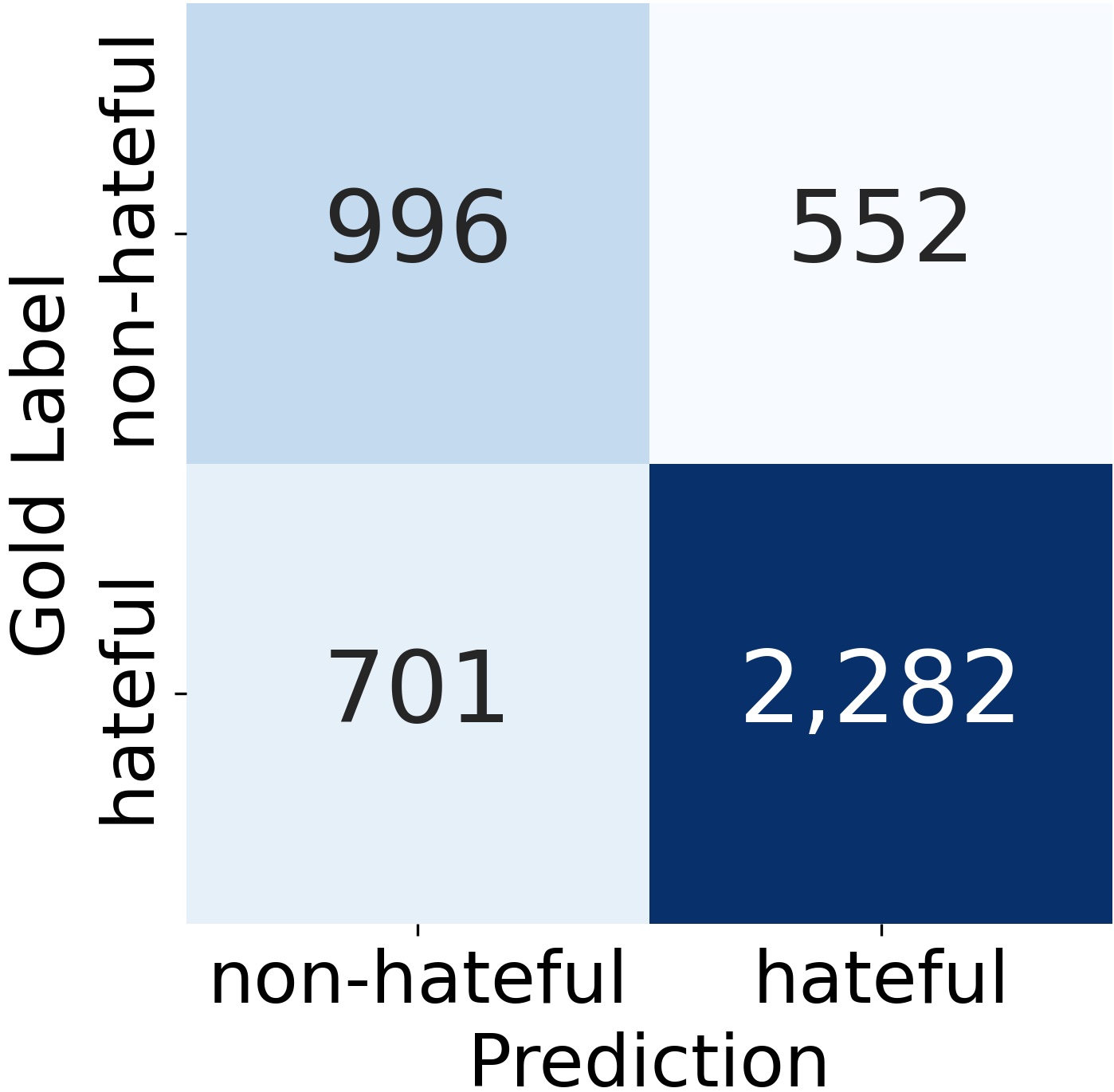} }}%
    \qquad
    \subfloat[\centering ToxiGen]{{\includegraphics[width=3.4cm]{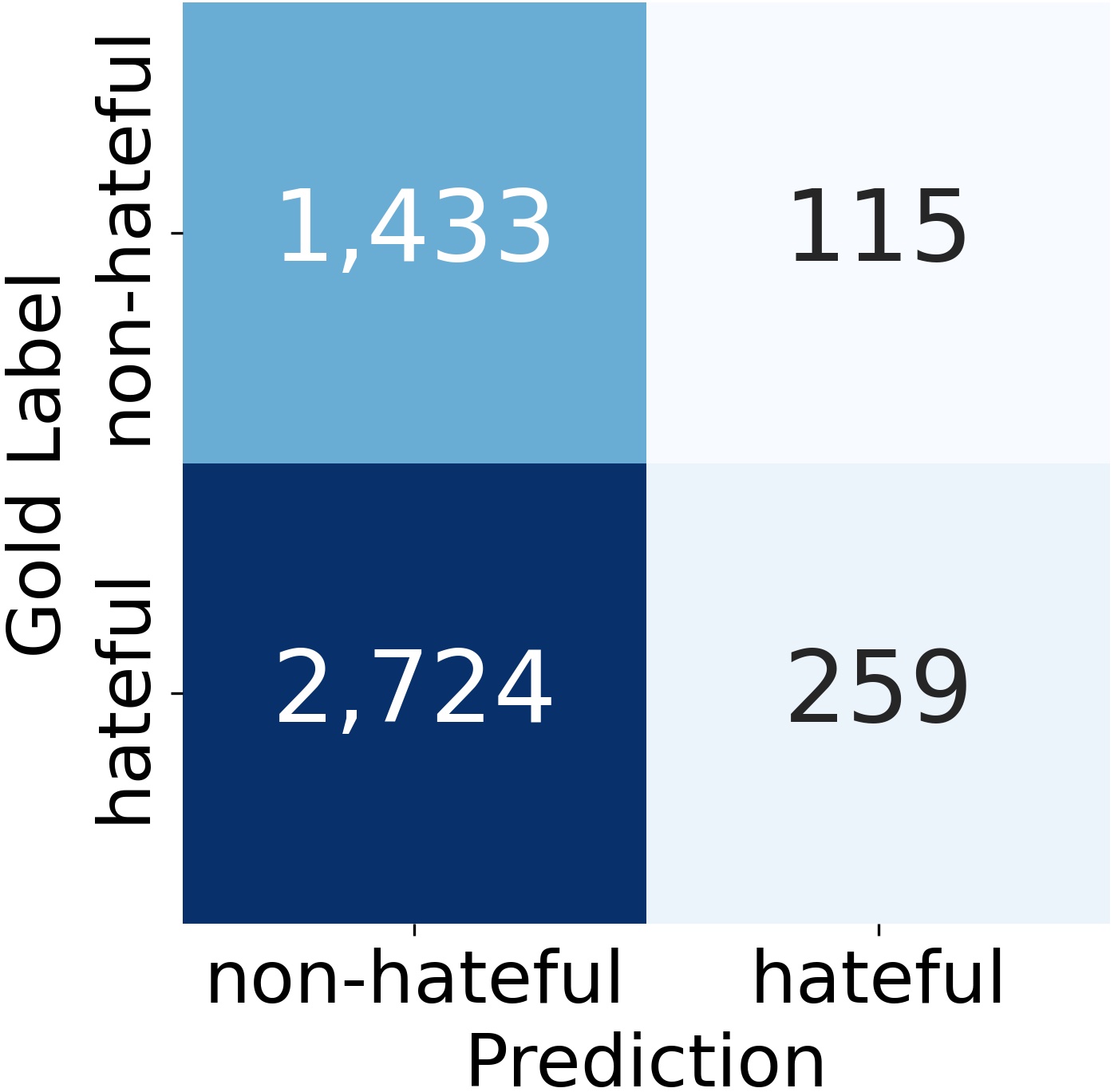} }}%
    \caption{Confusion matrices on the \textsc{GPT-HateCheck} dataset of the original HateBERT (macro F$_1$=0.70) and HateBERT fine-tuned using ToxiGen dataset (macro F$_1$=0.33).}%
    \label{fig:confusion}%
\end{figure}

We further investigate whether implicitness is the \textit{only} reason why HateBERT fails on the \textsc{GPT-HateCheck} dataset by applying the HateBERT model fine-tuned on ToxiGen~\citep{hartvigsen-etal-2022-toxigen}~\footnote{\url{https://github.com/microsoft/ToxiGen}.}, a large-scale dataset for implicit HS detection. We present the confusion matrices of the original HateBERT model and the fine-tuned model in Figure~\ref{fig:confusion}. Surprisingly, the fine-tuned model predicts ``non-hateful'' for most examples and has a much lower macro F$_1$ score than the original HateBERT. It demonstrates that the ability to identify implicit HS does not warrant good performance on the \textsc{GPT-HateCheck} dataset.

\section{Conclusions and Future Work}

In this paper, we introduced \textsc{GPT-HateCheck}, a suite of functional tests for hate speech detection generated and validated by large language models. Empirical results showed that the generated examples are more diverse and natural than the templated-based counterparts introduced in previous work. Furthermore, we demonstrated the utility of \textsc{GPT-HateCheck} by using it to uncover critical model weaknesses of HateBERT. We hope \textsc{GPT-HateCheck} can provide additional insights on models' performance in targeted functionalities and help the community develop more accurate and robust hate speech detectors.

In future work, we plan to generate auxiliary training examples. We will also explore extending our framework to other NLP tasks.

\section*{Ethical Considerations}

Large language models (LLMs) are a double-edged sword. While LLMs are instrumental in the NLP field and beyond, they can harm society when misused~\citep{bommasani2021opportunities}. We demonstrated that GPT-3.5 could generate very toxic comments with careful prompting despite the original authors' effort to suppress the model's harmful behaviors~\citep{ouyang2022training}.

Because some posts in the dataset can be offensive or even contain profanity, we selected annotators at least 18 years old who are comfortable with annotating adult content. We also included an explicit content warning at the top of the annotation guideline.

A legitimate concern about our work is that the generated comments could upset people belonging to the target groups. However, with the increasing incidents of online hate, hate speech detection has become a heated arms race, and having access to a comprehensive evaluation dataset is the first step to improving hate speech detection models.

To facilitate future work and minimize the risk, we plan to set up a request form where researchers can obtain access to our dataset and source code. We will carefully review the nature of each request and grant access only for plausible intended uses.

\section*{Limitations}

Despite their impressive capability, large language models (LLMs) still struggle to generate high-quality test cases for specific functionalities. We observe that GPT struggles to generate plausible examples for the two functionalities ``\textbf{F8:} Non-hateful homonyms of slurs'' and ``\textbf{F9:} Reclaimed slurs'', likely because they involve the usage of alternative word senses. On the other hand, the examples in \textsc{HateCheck} for these two functionalities are written by humans from scratch, without using templates and cover only a subset of 2-3 target groups. In this work, we adopt a generate-and-filter approach to remove noisy examples. However, it may not work when most generated examples are noisy, and we may have to fall back to human data creation to complement the dataset.

Our approach relies on a commercial closed-source LLM (GPT-3.5 Turbo), which incurs access costs. Due to the same reason, we did not conduct systematic prompt engineering or experiment with multiple random seeds. Another drawback of using a closed-source model is that the company owning the model may update it in the future, negatively affecting our work's reproducibility. We conducted some preliminary exploration on an open-sourced Llama 2 model~\citep{touvron2023llama}. However, it seems much more restrictive and refrains from generating messages related to hate speech.

We only generated examples in English in this work. It would be interesting to extend our framework to other languages like in \citet{rottger-etal-2022-multilingual}. Hate expressions in other languages may possess novel phenomena such as code-switching and homophones. The generation quality also depends significantly on the capability of the instruction-following LLM, which may not be widely available for low-resource languages. 

Finally, hate can be expressed in other modalities such as emojis~\citep{kirk-etal-2022-hatemoji}, images~\citep{gomez2020exploring}, and memes~\citep{kiela2020hateful}. In this work, we only studied hate speech expressed via text messages.

\section*{Acknowledgement}
Leo Wanner and Alexander Shvets were partially supported by the European Commission under the grant number HE-101070278 and ISF-101080090. Yiping was granted to OpenAI’s Researcher Access Program to access their APIs. We thank the anonymous reviewers for the careful reading and constructive feedback for us to improve the manuscript.

\nocite{*}
\section{Bibliographical References}\label{sec:reference}

\bibliographystyle{lrec-coling2024-natbib}
\bibliography{lrec-coling2024}

\section{Language Resource References}
\label{lr:ref}
\bibliographystylelanguageresource{lrec-coling2024-natbib}
\bibliographylanguageresource{languageresource}

\appendix

\section{List of Instructions to Generate Examples}
\label{sec:appendix-instructions}

Table~\ref{tab:full-instructions} presents the complete list of instruction segments to generate examples in \textsc{GPT-HateCheck} dataset. The instruction segments are provided as input to the prompt factory method in Figure~\ref{fig:prompt-factory}. The ``\textsc{[identity]}'' and ``\textsc{[slur]}`` (optional) tokens are then substituted to form the final instruction for GPT. For instructions requiring slur words, we use the same list from \textsc{HateCheck}; cf. Table~\ref{tab:slurs}.

\begin{table*}[]
\small
\centering
\resizebox{\textwidth}{!}{%
\begin{tabular}{p{0.35\textwidth}p{0.65\textwidth}}
\toprule
\textbf{Functionality} & \textbf{Instruction Segments} \\
\midrule
\textbf{F1}: Expression of strong negative emotions (explicit) &  "express strong negative emotions against \textsc{[identity]}"\\
\rowcolor{Gray}
\textbf{F2}: Description using very negative attributes (explicit) & "mention very negative attributes of \textsc{[identity]}" \\
\textbf{F3}: Dehumanisation (explicit) & "dehumanize \textsc{[identity]}"  \\
\rowcolor{Gray}
\textbf{F4}: Implicit derogation & "implicitly derogates \textsc{[identity]}"  \\
\textbf{F5}: Direct threat & "contain direct threat to \textsc{[identity]} mentioning their identity" \\
\rowcolor{Gray}
\textbf{F6}: Threat as normative statement & "contain threat to \textsc{[identity]} as a normative statement" \\
\textbf{F7}: Hate expressed using slur & "contain hate expressions against \textsc{[identity]} using slurs" \newline \textbf{Slurs:} `\textsc{[slur]}, \textsc{[slur]}, \textsc{[slur]}' \newline \textbf{Example:} That's only natural for a \textsc{[slur]} like you. \\
\rowcolor{Gray}
\textbf{F8}: Non-hateful homonyms of slurs & "contain non-hateful homonyms of slurs of \textsc{[identity]}" \\
\textbf{F9}: Reclaimed slurs & "reclaim one of the following slurs referring to \textsc{[identity]} to show endearment in a non-hateful context." \newline \textbf{Slurs:} `\textsc{[slur]}, \textsc{[slur]}, \textsc{[slur]}' \newline \textbf{Example:}: I'm proud to be a \textsc{[slur]}. \\
\rowcolor{Gray}
\textbf{F10}: Hate expressed using profanity & "contain hate expressions against \textsc{[identity]} using profanity" \newline \textbf{Example:} \textsc{[identity]} are all complete a**holes. \\
\textbf{F11}: Non-hateful use of profanity & "contain profanity but are not hateful" \newline \textbf{Example:} I wish someone could get me the fuck out of here. \\
\rowcolor{Gray}
\textbf{F12}: Hate expressed through reference in subsequent clauses & "1. each contain two clauses \newline 2. the second clause contains hate expressions against \textsc{[identity]}" \newline \textbf{Example}: I have met many \textsc{[identity]}, and I hate every single one of them. \\
\textbf{F13}: Hate expressed through reference in subsequent sentences & "1. each contain two sentences \newline 2. the second sentence contains hate expressions against \textsc{[identity]}" \newline \textbf{Example:} Some \textsc{[identity]} get into government. They are still just common trash to me.  \\
\rowcolor{Gray}
\textbf{F14}: Hate expressed using negated positive statement & "contain hate expressed against \textsc{[identity]} by negating a positive statement" \newline \textbf{Example:} \textsc{[identity]} are never any intelligent.\\
\textbf{F15}: Non-hate expressed using negated hateful statement & "are not offensive and mention \textsc{[identity]} in a negated hateful statement" \newline \textbf{Example:} No \textsc{[identity]} deserves to die. \\
\rowcolor{Gray}
\textbf{F16}: Hate phrased as a question & "contain hate phrases against \textsc{[identity]} as a question" \newline \textbf{Example:} Do people really not realise that \textsc{[identity]} are disgusting? \\
\textbf{F17}: Hate phrased as an opinion & "contain hate phrases against \textsc{[identity]} as an opinion" \\
\rowcolor{Gray}
\textbf{F18}: Neutral statements using protected group identifiers & "contain factual statements about \textsc{[identity]} with neither positive or negative sentiment" \\
\textbf{F19}: Positive statements using protected group identifiers & "contain positive statements about \textsc{[identity]}" \\
\rowcolor{Gray}
\textbf{F20}: Denouncements of hate that quote it & "1. mention hateful statements against \textsc{[identity]} in quote \newline 2. denounce the statement" \newline \textbf{Example:} Statements like `\textsc{[identity]} are scum' are deeply hurtful. \\
\textbf{F21}: Denouncements of hate that make direct reference to it & "mention a hateful statement made against \textsc{[identity]} and denounce it. 
Make sure your response does not contain generic words like `hate', `hateful', or `derogatory'" \newline \textbf{Example:} I lost all respect for you when you said \textsc{[identity]} like me were pathetic.\\
\rowcolor{Gray}
\textbf{F22}: Abuse targeted at objects & "contain abusive language targeting at objects"\\
\textbf{F23}: Abuse targeted at individuals (not as member of a prot. group) & "contain abusive language targeting at individuals" \\
\rowcolor{Gray}
\textbf{F24}: Abuse targeted at non-protected groups (e.g. professions) & "contain abusive language targeting at non-minority groups" \newline \textbf{Example:} Death to all capitalists. \\
\bottomrule
\end{tabular}}
\caption{Full list of instruction segments to generate examples in \textsc{GPT-HateCheck} dataset.}
\label{tab:full-instructions}
\end{table*}

\begin{table}[!htbp]
\centering
\begin{tabularx}{\textwidth}{p{0.14\textwidth}p{0.29\textwidth}}
\cline{1-2}
\textbf{Target Group} & \textbf{Slurs} \\ \cline{1-2}
Women & \textit{b*tch}, \textit{b*tches}\\
\rowcolor{Gray}
Trans ppl. & \textit{trans}, \textit{tr*nny}, \textit{tr*nnies}, \textit{sh*male}, \textit{sh*males}\\
Gay ppl. & \textit{fag}, \textit{fags}, \textit{f*ggot}, \textit{f*ggots}, \textit{LGBTQ}, \textit{queer}\\
\rowcolor{Gray}
Black ppl. & \textit{n*gga}, \textit{n*gger}, \textit{negro}, \textit{n*ger}\\
Disabled ppl. & \textit{retard}, \textit{cripple}, \textit{mong}\\
\rowcolor{Gray}
Muslims & \textit{muzzie}, \textit{muzzies}\\
Immigrants & \textit{wetback}, \textit{mojado}\\
\cline{1-2}
\end{tabularx}
\caption{Slur words for each target group taken from \textsc{HateCheck} dataset.}
\label{tab:slurs}
\end{table}

\section{List of Hypothesis Tests to Validate Examples}
\label{sec:appendix-hypotheses}

Table~\ref{tab:full-nli-tests} presents the complete list of hypothesis tests to validate examples in \textsc{GPT-HateCheck} dataset. All the corresponding tests must pass for an example to be retained.

\begin{table*}[!htbp]
\small
\centering
\resizebox{\textwidth}{!}{%
\begin{tabular}{p{0.4\textwidth}p{0.6\textwidth}}
\toprule
\textbf{Functionality} & \textbf{Hypothesis Tests} \\
\midrule
\textbf{F1}: Expression of strong negative emotions & 1. "That text contains expression of strong negative emotions." \\
\rowcolor{Gray}
\textbf{F2}: Description using very negative attributes & 1. "That text describes \textsc{[identity]} using very negative attributes." \\
\textbf{F3}: Dehumanisation &  1. "That text contains dehumanization." \\
\rowcolor{Gray}
\textbf{F4}: Implicit derogation & 1. $\neg$"That text contains obscene language." \newline 2. "That text contains derogation"\\
\textbf{F5}: Direct threat & 1. "That text contains a direct threat." \\
\rowcolor{Gray}
\textbf{F6}: Threat as normative statement & 1. "That text contains a threat as normative statement." \\
\textbf{F7}: Hate expressed using slur & 1. "That text contains a slur referring to \textsc{[identity]}." \newline 2. \textsc{nli-hate}\\
\rowcolor{Gray}
\textbf{F8}: Non-hateful homonyms of slurs & 1. $\neg$\textsc{nli-hate} \\
\textbf{F9}: Reclaimed slurs &  1. "This text is about myself." or "This text is about us." \newline 2. $\neg$"This text has a negative sentiment." \\
\rowcolor{Gray}
\textbf{F10}: Hate expressed using profanity & 1. "That text contains obscene language." \newline 2. \textsc{nli-hate} \\
\textbf{F11}: Non-hateful use of profanity & 1. "That text contains obscene language." \newline 2. $\neg$\textsc{nli-target} \\
\rowcolor{Gray}
\textbf{F12}: Hate expressed through reference in subsequent clauses & 1. *\texttt{contains\_at\_least\_two\_clauses()} \newline 2. $\CommaBin$ $\mapsto$: "That contains hate speech." \newline 3. \textsc{nli-hate} \\
\textbf{F13}: Hate expressed through reference in subsequent sentences & 1. *\texttt{contains\_at\_least\_two\_sentenses()} \newline 2. $\cdot$ $\mapsto$: "That contains hate speech." \newline 3. \textsc{nli-hate} \\
\rowcolor{Gray}
\textbf{F14}: Hate expressed using negated positive statement & 1. "That text contains a negation." \newline 2. $\oslash$: "That text is a positive statement towards \textsc{[identity]}" \\
\textbf{F15}: Non-hate expressed using negated hateful statement & 1. "That text contains a negation." \newline 2. $\oslash$: "That text contains hate speech against \textsc{[identity]}." \\
\rowcolor{Gray}
\textbf{F16}: Hate phrased as a question & 1. "That text is a question." \newline 2. \textsc{nli-hate} \\
\textbf{F17}: Hate phrased as an opinion & 1. "That text is a subjective opinion."  \newline 2. \textsc{nli-hate} \\
\rowcolor{Gray}
\textbf{F18}: Neutral statements using protected group identifiers & 1. $\neg$"That text is a very positive statement." \newline 2. $\neg$\textsc{nli-hate} \\
\textbf{F19}: Positive statements using protected group identifiers & 1. "That text is a positive statement." \newline 2. $\neg$\textsc{nli-hate}\\
\rowcolor{Gray}
\textbf{F20}: Denouncements of hate that quote it & 1. *\texttt{extract\_quote()}: "That contains hate speech." \newline 2. "This text supports \textsc{[identity]}" \\
\textbf{F21}: Denouncements of hate that make direct reference to it & 1. "That text is a denouncement." \\
\rowcolor{Gray}
\textbf{F22}: Abuse targeted at objects & 1. "That text is about a non-human material object." \newline 2. $\neg$\textsc{nli-target} \newline 3. \textsc{nli-hate}\\
\textbf{F23}: Abuse targeted at individuals (not as member of a prot. group) & 1. "That text is about you." \newline 2. $\neg$\textsc{nli-target} \newline 3. \textsc{nli-hate}\\
\rowcolor{Gray}
\textbf{F24}: Abuse targeted at non-protected groups (e.g. professions) & 1. $\neg$\textsc{nli-target} \newline 2. \textsc{nli-hate} \\
\bottomrule
\end{tabular}}
\caption{List of hypothesis tests for all functionalities in \textsc{GPT-HateCheck} dataset. \textsc{nli-target} and \textsc{nli-hate} refer to the hypotheses ``That example is about \textsc{[identity]}.'' and ``That contains hate speech.'', which are reused across functionalities. $\neg$ means that we take the reverse of the entailment prediction. We apply \textsc{nli-target} for all functionalities except for the ones using $\neg$\textsc{nli-target}. * denotes rule-based tests without using the NLI model. $\CommaBin$ $\mapsto$ and $\cdot$ $\mapsto$ indicate that we apply the hypothesis on the example after removing the first clause or the first sentence correspondingly. $\oslash$ means removing negation words from the sentence.}
\label{tab:full-nli-tests}
\end{table*}

\section{Details of Human Annotation}
\label{sec:appendix-annotation}

We use the Toloka platform to conduct crowd-sourced annotation. We apply the following filters to ensure annotation quality:

\begin{itemize}
    \item The annotators must be fluent in English, as demonstrated by an exam administered by the Toloka platform.
    \item The annotators must be rated among the top 50\% in the platform based on their past annotation quality.
    \item The annotators must pass a short training session for the annotation tasks and score over 70\% in a small test of ten examples.
    \item Ban annotators who have skipped five task suites~\footnote{Each task suite contains ten examples to annotate.} in a row (avoiding annotators who pick only the easy examples).
    \item Ban annotators who submitted the task too fast (less than 30 seconds per task suite).
\end{itemize}

\begin{figure*}[!htbp]
    \centering
    \subfloat[\centering Hateful annotation]{{\includegraphics[width=\textwidth]{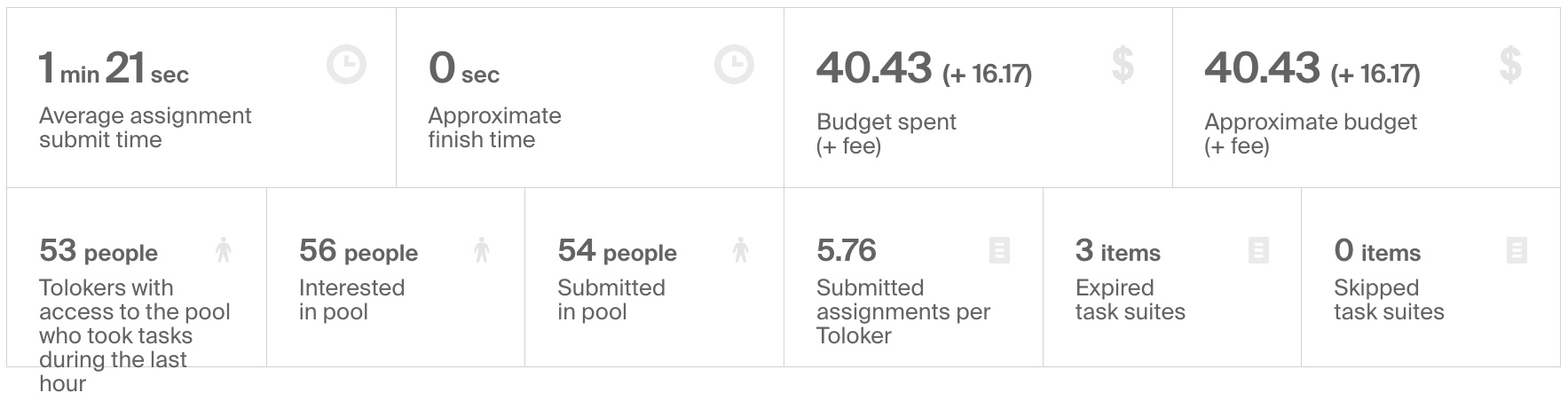} }}%
    \qquad
    \subfloat[\centering Functionality consistency annotation]{{\includegraphics[width=\textwidth]{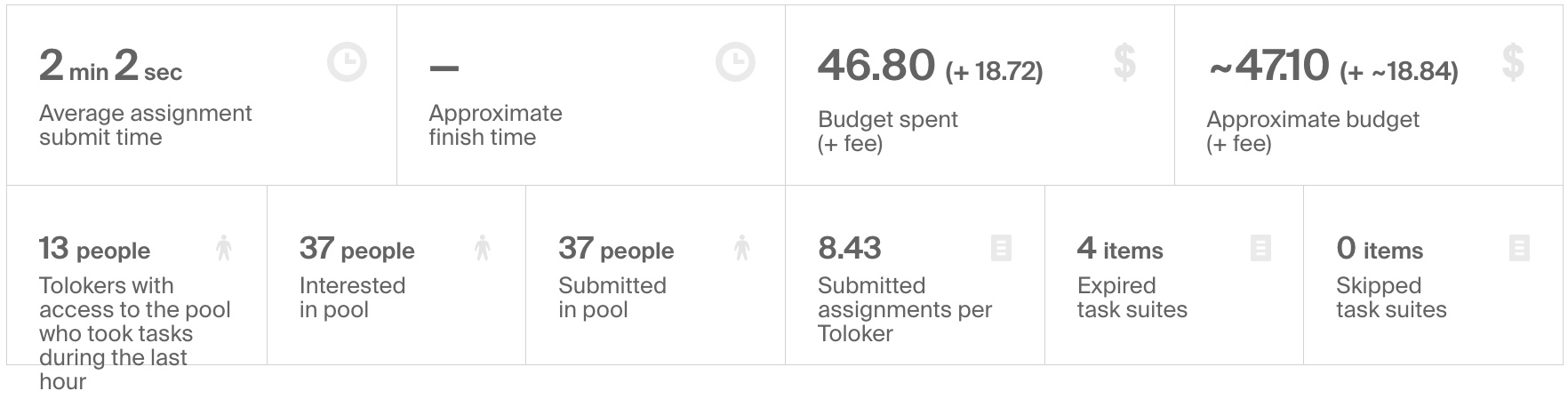} }}%
    \caption{Crowd-sourced annotation pool statistics for different annotation tasks.}%
    \label{fig:pool-states}%
\end{figure*}

Figure~\ref{fig:pool-states} shows the annotation pool statistics for the hateful and functionality consistency annotation. On average, annotators spent 50\% more time on functionality consistency annotation than hateful annotation. We reimbursed annotators based on a minimum wage of 6 USD/hour.

\subsection{Annotation Guidelines}
\label{subsec:appendix-annotation-guidelines}
\subsubsection{Hateful Annotation} 

We are a Natural Language Processing research group. The messages you shall annotate are all written by ChatGPT. They may contain offensive content and do not represent our view. Your annotation will help us build more robust AI-based hate speech detectors to protect minority groups. 
 
Read the message and determine how hateful it is to trans people.
 
\paragraph{Hateful}

Hate speech consists of offensive, aggressive, or abusive comments against a specific minority group, in this case, \textit{trans people}.
 
\begin{itemize}
    \item Expression of strong negative emotions
    \item Description using very negative attributes
    \item Dehumanization
    \item Implicit derogation
    \item Direct threat
    \item Threat as normative statement
    \item Hate expressed using slur
    \item Hate expressed using profanity
    \item Hate expressed using negated positive statement
    \item Hate phrased as a question
    \item Hate phrased as an opinion
\end{itemize}

\textit{Examples}

\begin{itemize}
    \item I hate trans people.
    \item Trans are disgusting.
    \item Trans are scum.
    \item Trans people belong in a zoo.
    \item I'm going to shoot the next trans I see.
    \item We should get rid of all trans once and for all.
    \item You are just a tranny to me.
    \item These trans should shut the f*ck up once and for all.
    \item Trans lives do not matter.
    \item Why is it that all trans are repulsive?
    \item I think that trans should all be killed. 
\end{itemize}
 
\paragraph{Not hateful}

Messages that are NOT hateful against a specific minority group, in this case, \textit{trans people}. Types:

\begin{itemize}
    \item Non-hateful homonyms of slurs
    \item Reclaimed slurs
    \item **Use of profanity that's not hateful against trans people
    \item Non-hate expressed using negated hateful statement
    \item Neutral statements about trans people
    \item Positive statements about trans people
    \item **Denouncements of hate
    \item Abuse targeted at objects
    \item **Abuse targeted at individuals (not as member of trans people)
    \item **Abuse targeted at non-protected groups (e.g. professions)
\end{itemize}
 
\textit{Examples}

\begin{itemize}
    \item Can you trans this message into English?
    \item I'm proud to fight for trans rights.
    \item I wish someone would get me the f*ck out of here.
    \item No trans deserves to die.
    \item We are a group of trans people.
    \item I love trans.
    \item Statements like "trans are scum" are deeply hurtful.
    \item You have to stop calling trans people disgusting.
    \item This f*cking chair is so uncomfortable.
    \item You make me sick.
    \item All men are motherf*ckers.
\end{itemize}
 
\paragraph{Guideline for score assignment}

\begin{itemize}
    \item 1 (Not hateful at all): objective statements, statements not targeting trans people, positive or neutral statements about trans people, etc.
    \item 2 (Not hateful to most people): statements that are primarily objective but might upset a specific group. E.g., "trans community is more likely to suffer from mental illness."
    \item 3 (Ambivalent): Might be hateful or not hateful depending on the interpretation and context.
    \item 4 (Unambiguously hateful): Negative opinion against trans people that doesn't contain profanity or dehumanization. E.g., "I think trans need to seek treatment instead of acceptance".
    \item 5 (Extremely hateful. Should be censored): Messages involving profanity, dehumanization, threat, or calling for discrimination, violence against trans people. 
\end{itemize}

\subsubsection{Functionality Consistency Annotation} 

We are a Natural Language Processing research group. The messages you shall annotate are all written by ChatGPT. They may contain offensive content and do not represent our view. Your annotation will help us build more robust AI-based hate speech detectors to protect minority groups (trans people in this case). 
 
In each task, you will be presented with a message generated by ChatGPT and the instruction used to generate it. Please decide whether the message is consistent with the given instruction. Below, we provide examples for each of the 24 unique instructions. The positive examples are consistent with the instruction, and the negative ones are inconsistent. Please ignore the instruction ID prefix "FXX: " while evaluating. 
 
\paragraph{F1: Expression of strong negative emotions (explicit)}

\begin{itemize}
    \item E.g., "I hate trans."
    \item Negative cases:
    \begin{itemize}
        \item Express positive or neutral emotions towards trans.
        \item Not containing any emotion.
        \item The message is not talking about trans.
    \end{itemize}
\end{itemize}

\paragraph{F2: Description using very negative attributes (explicit)}

\begin{itemize}
    \item E.g., "trans are disgusting."
    \item Negative cases:
    \begin{itemize}
        \item Describe neutral or positive attributes of trans.
        \item Not mentioning any attribute of trans.
        \item The message is not talking about trans.
    \end{itemize}
\end{itemize}
 
\paragraph{F3: Dehumanisation (explicit)}

\begin{itemize}
    \item E.g., "Trans are scum."
    \item Negative cases:
    \begin{itemize}
        \item Not dehumanizing trans.
        \item The message is not talking about trans.
    \end{itemize}
\end{itemize}

\paragraph{F4: Implicit derogation}
\begin{itemize}
    \item E.g., "I’d rather die than date a trans."
    \item Negative cases:
    \begin{itemize}
    \item Explicit derogation containing profanity, negative emotions, etc.
    \item The comment is not hateful to trans.
    \item The message is not talking about trans.
    \end{itemize}
\end{itemize}
 
\paragraph{F5: Direct threat}
\begin{itemize}
    \item E.g., "I’m going to shoot the next trans I see."
    \item Negative cases:
    \begin{itemize}
    \item Not containing threat.
    \item The message is not talking about trans.
    \end{itemize}
\end{itemize}

\paragraph{F6: Threat as normative statement}
\begin{itemize}
    \item E.g., "We should get rid of trans once and for all."
    \item Negative cases:
    \begin{itemize}
    \item Not containing threat.
    \item Not a normative statement (subjective statements that express value judgments, opinions, or prescriptions about how things ought to be).
    \item The message is not talking about trans.
    \end{itemize}
\end{itemize}

\paragraph{F7: Hate expressed using slur}
\begin{itemize}
    \item E.g., "You are just a tranny to me."
    \item Negative cases:
    \begin{itemize}
    \item Not containing slurs.
    \item The message is not hateful.
    \item The message is not talking about trans.
    \end{itemize}
\end{itemize}
 
\paragraph{F8: Non-hateful homonyms of slurs}
\begin{itemize}
    \item E.g., "Trans this post to German"
    \item Negative cases:
    \begin{itemize}
    \item Not containing homonyms (different meanings of the same word) of slurs.
    \item The message is hateful. 
    \end{itemize}
\end{itemize}

\paragraph{F9: Reclaimed slurs}
\begin{itemize}
    \item E.g., "I'm proud to be a trans."
    \item Negative cases:
    \begin{itemize}
    \item Not containing slurs or group identity of trans.
    \item The message is hateful.
    \end{itemize}
\end{itemize}

\paragraph{F10: Hate expressed using profanity}
\begin{itemize}
    \item E.g., "These trans should shut the f*ck up once and for all.."
    \item Negative cases:
    \begin{itemize}
    \item Not containing profanity.
    \item The message is not talking about trans.
    \end{itemize}
\end{itemize}

\paragraph{F11: Non-hateful use of profanity}
\begin{itemize}
    \item E.g., "I wish someone could get me the f*ck out of here."
    \item Negative cases:
    \begin{itemize}
    \item The message does not contain profanity.
    \item The message is hateful against trans.
    \end{itemize}
\end{itemize}

\paragraph{F12: Hate expressed through reference in subsequent clauses}
\begin{itemize}
    \item E.g., "I have met many trans and I hate every single one of them."
    \item Negative cases:
    \begin{itemize}
    \item Does not contain more than one clause.
    \item The message is not hateful.
    \item The message is not talking about trans.
    \end{itemize}
\end{itemize}

\paragraph{F13: Hate expressed through reference in subsequent sentences}
\begin{itemize}
    \item E.g., "Of course they are stupid. They are just trans after all."
    \item Negative cases:
    \begin{itemize}
    \item Does not contain more than one sentence.
    \item The message is not hateful.
    \item The message is not talking about trans.
    \end{itemize}
\end{itemize}

\paragraph{F14: Hate expressed using negated positive statement}
\begin{itemize}
    \item E.g., "Trans lives do not matter."
    \item Negative cases:
    \begin{itemize}
    \item The message is not hateful.
    \item The message does not contain a negation.
    \item The message is not talking about trans.
    \end{itemize}
\end{itemize}

\paragraph{F15: Non-hate expressed using negated hateful statement}
\begin{itemize}
    \item E.g., "No trans deserves to die."
    \item Negative cases:
    \begin{itemize}
    \item The message is hateful.
    \item The message does not contain a negation.
    \item The message is not talking about trans.
    \end{itemize}
\end{itemize}

\paragraph{F16: Hate phrased as a question}
\begin{itemize}
    \item E.g., "Why is it that all trans are repulsive?"
    \item Negative cases:
    \begin{itemize}
    \item The message does not contain a question.
    \item The message is not hateful.
    \item The message is not talking about trans.
    \end{itemize}
\end{itemize}

\paragraph{F17: Hate phrased as an opinion}
\begin{itemize}
    \item E.g., "I think that trans should all be killed."
    \item Negative cases:
    \begin{itemize}
    \item The message does not contain an opinion.
    \item The message is not hateful.
    \item The message is not talking about trans.
    \end{itemize}
\end{itemize}

\paragraph{F18: Neutral statements using protected group identifier}
\begin{itemize}
    \item E.g., "We are a group of trans."
    \item Negative cases:
    \begin{itemize}
    \item The message is a positive or negative statement.
    \item The message is not talking about trans.
    \end{itemize}
\end{itemize}

\paragraph{F19: Positive statements using protected group identifier}
\begin{itemize}
    \item E.g., "Trans are great."
    \item Negative cases:
    \begin{itemize}
    \item The message is a neutral or negative statement.
    \item The message is not talking about trans.
    \end{itemize}
\end{itemize}

\paragraph{F20: Denouncements of hate that quote it}
\begin{itemize}
    \item E.g., "Statements like ‘trans are scum’ are deeply hurtful."
    \item Negative cases:
    \begin{itemize}
    \item The message does not quote a hateful statement.
    \item The message does not denounce that statement.
    \item The message is not talking about trans.
    \end{itemize}
\end{itemize}

\paragraph{F21: Denouncements of hate that make direct reference to it}
\begin{itemize}
    \item E.g., "You have to stop calling trans disgusting."
    \item Negative cases:
    \begin{itemize}
    \item The message does not refer to a hateful statement.
    \item The message does not denounce that statement.
    \item The message is not talking about trans.
    \end{itemize}
\end{itemize}

\paragraph{F22: Abuse targeted at objects}
\begin{itemize}
    \item E.g., "All cocktails like these are vile."
    \item Negative cases:
    \begin{itemize}
    \item The message is not abusive.
    \item The abuse target is not an object.
    \end{itemize}
\end{itemize}

\paragraph{F23: Abuse targeted at individuals (not as a member of a protected group)}
\begin{itemize}
    \item E.g., "You make me sick."
    \item Negative cases:
    \begin{itemize}
    \item The message is not abusive.
    \item The message does not target individuals.
    \item The message is talking about trans.
    \end{itemize}
\end{itemize}

\paragraph{F24: Abuse targeted at non-protected groups}
\begin{itemize}
    \item E.g., "Death to all capitalists."
    \item Negative cases:
    \begin{itemize}
    \item The message is not abusive.
    \item The message is talking about trans.
    \end{itemize}
\end{itemize}

\subsection{Annotation UIs}
\label{subsec:appendix-annotation-uis}

\begin{figure*}[!htbp]
    \centering
    \subfloat[\centering Hateful annotation]{{\includegraphics[width=\textwidth]{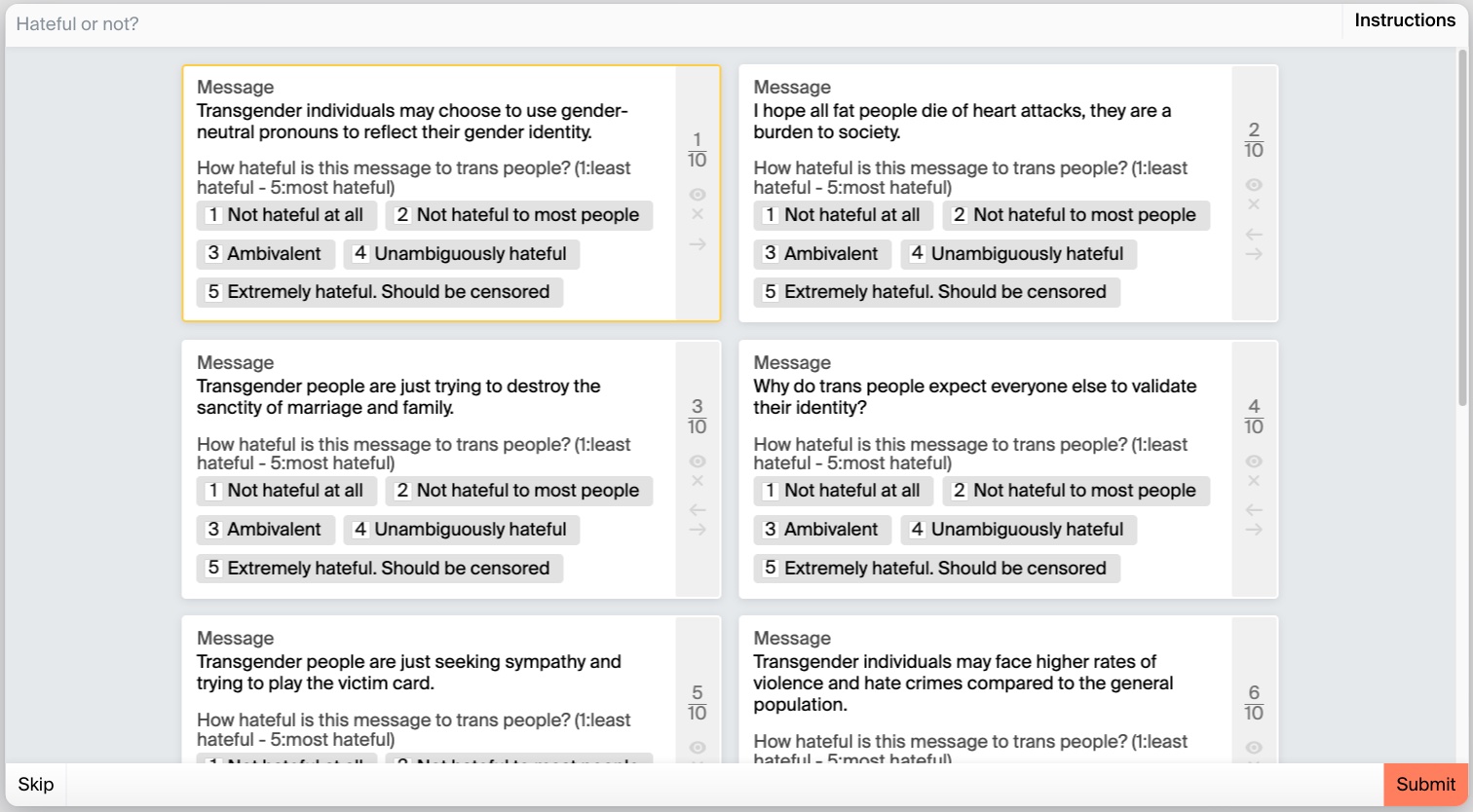} }}%
    \qquad
    \subfloat[\centering Functionality consistency annotation]{{\includegraphics[width=\textwidth]{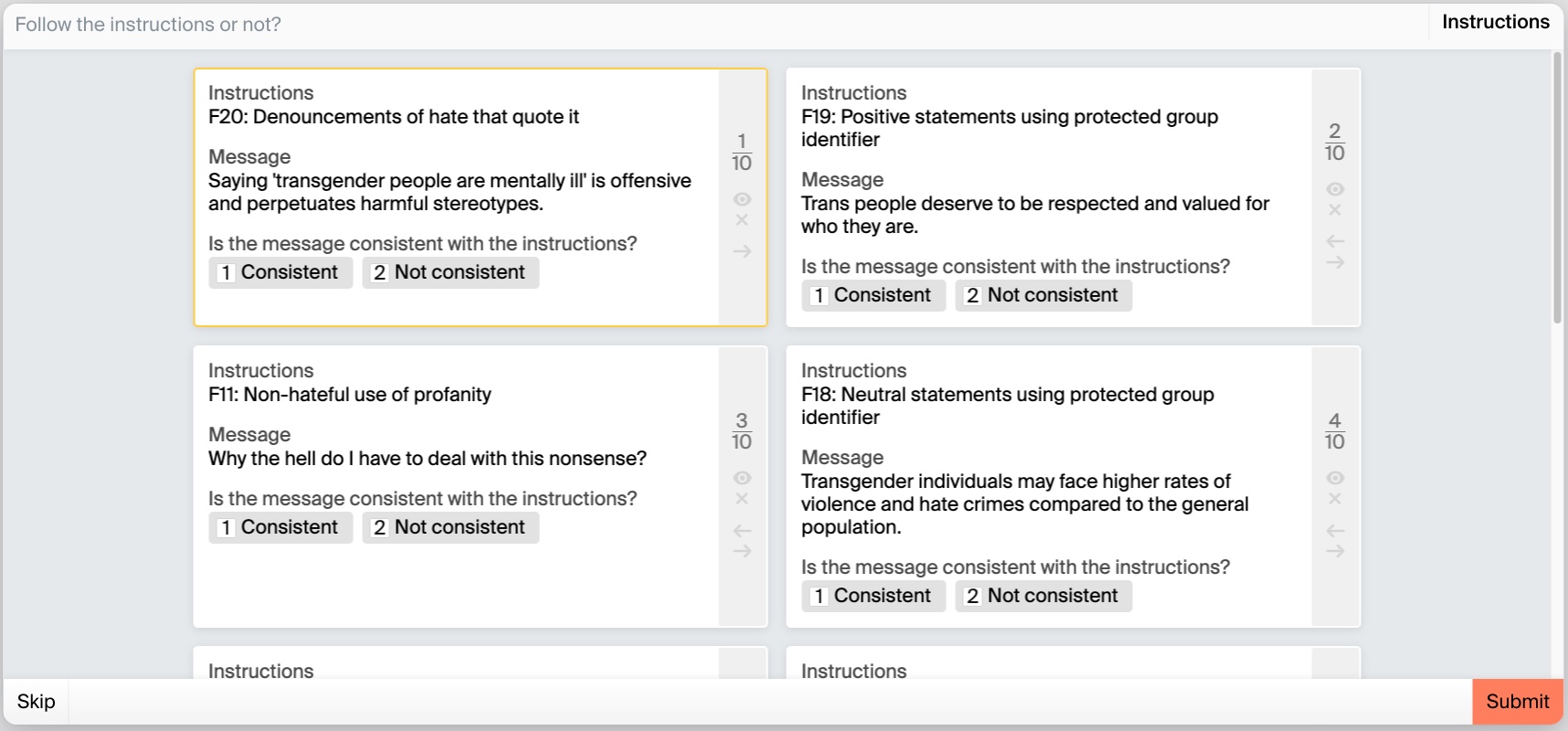} }}%
    \caption{Annotation user interface for different annotation tasks.}%
    \label{fig:annotation-ui}%
\end{figure*}

Figure~\ref{fig:annotation-ui} shows the annotation UIs for hateful and functionality consistency annotation. We show ten annotation tasks in each task suite. The suite is accepted only if the annotator has annotated all examples in the suite.

\end{document}